\documentclass[a4paper,twoside]{article}

\usepackage{epsfig}
\usepackage{subcaption}
\usepackage{calc}
\usepackage{amssymb}
\usepackage{amstext}
\usepackage{amsmath}
\usepackage{amsthm}
\usepackage{multicol}
\usepackage{pslatex}
\usepackage{apalike}

\usepackage{algorithm,algorithmic}

\usepackage{SCITEPRESS}     

\begin{document}

\title{Latent Space Conditioning on Generative Adversarial Networks}

\author{\authorname{Ricard Durall\sup{1,2,3}, Kalun Ho\sup{1,3,4}, Franz-Josef Pfreundt\sup{1} and Janis Keuper\sup{1,5}}
\affiliation{\sup{1}Fraunhofer ITWM, Germany}
\affiliation{\sup{2}IWR, University of Heidelberg, Germany}
\affiliation{\sup{3}Fraunhofer Center Machine Learning, Germany}
\affiliation{\sup{4}Data and Web Science Group, University of Mannheim, Germany}
\affiliation{\sup{5}Institute for Machine Learning and Analytics, Offenburg University, Germany}
}

\keywords{Generative Adversarial Network, Unsupervised Conditional Training, Representation Learning.}

\abstract{ Generative adversarial networks are the state of the art approach towards learned synthetic image generation. Although early successes were mostly unsupervised, bit by bit, this trend has been superseded by approaches based on labelled data. These supervised methods allow a much finer-grained control of the output image, offering more flexibility and stability.  Nevertheless, the main drawback of such models is the necessity of annotated data. In this work, we introduce an novel framework that benefits from two popular learning techniques, adversarial training and representation learning, and takes a step towards unsupervised conditional GANs. In particular, our approach exploits the structure of a latent space (learned by the representation learning) and employs it to condition the generative model. In this way, we break the traditional dependency between condition and label, substituting the latter by unsupervised features coming from the latent space. Finally, we show that this new technique is able to produce samples on demand keeping the quality of its supervised counterpart. }

\onecolumn \maketitle \normalsize \setcounter{footnote}{0} \vfill

\section{\uppercase{Introduction}}
\label{sec:introduction}

\begin{figure*}[t!]
\begin{subfigure}{\linewidth}
	\centering
	\includegraphics[width=\linewidth]{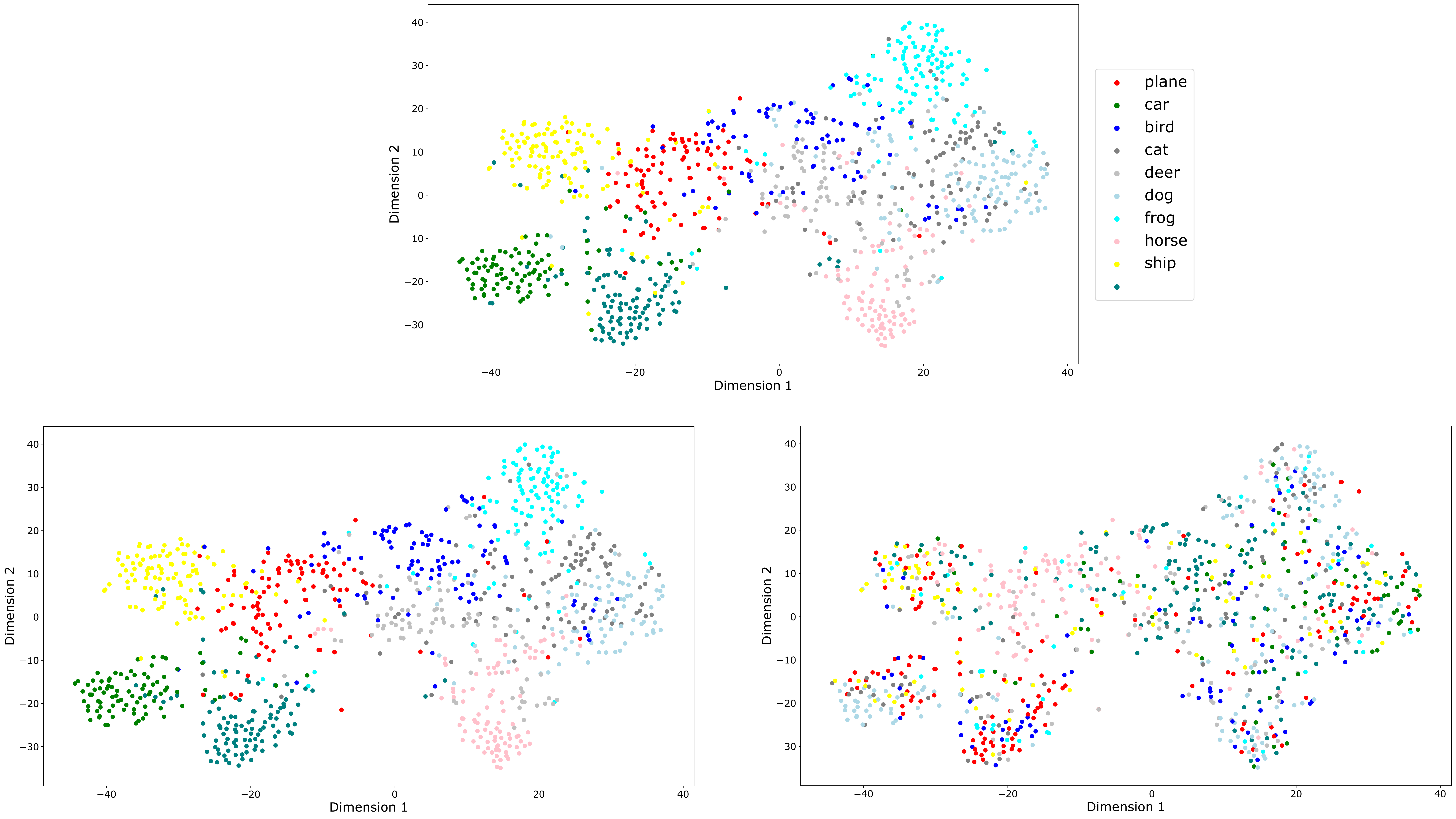}
	\caption{t-SNE visualizations. Upper-Center: Original latent space. Bottom-Left: Latent space according to our approach. Bottom-Right: Latent space according to standard approach\protect\footnotemark. Both bottom spaces represent the classes of the generated images given a latent code. Therefore, they should be as similar as possible to the original latent space.}
	\label{fig:tsne_cifar}
\end{subfigure}\\[2ex]
\begin{subfigure}{\linewidth}
	\centering
	\includegraphics[width=\linewidth]{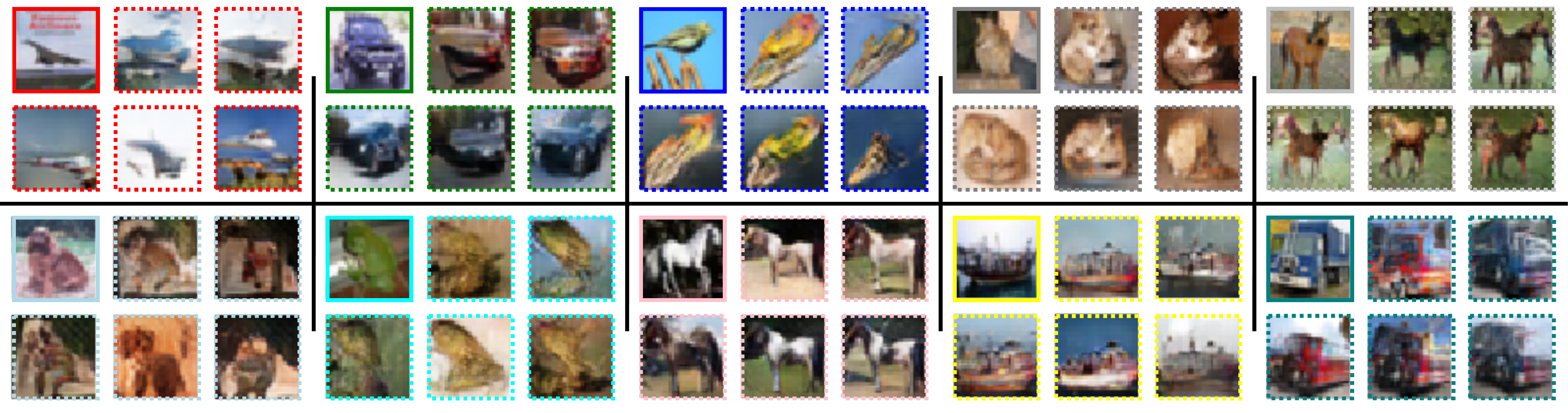}
	\caption{Random sets of generated samples trained on CIFAR10. The solid frames contain real images and the dashed the generated.}
	\label{fig:tsne_cifar2}
\end{subfigure}
\caption{Given a latent space, our approach exploits the structure using its features to condition the generative model. In this way, our system eventually can produce samples on demand. The code color is consistent within the whole figure.}
\label{fig:teaser}
\end{figure*}

\noindent  Generative Adversarial Networks (GANs) \cite{goodfellow2014generative} are one of the most prominent unsupervised generative models. Their framework involves training a generator and discriminator model in an adversarial game, such that the generator learns to produce samples from the data distribution. Training GANs is challenging because they require to deal with a minimax loss that needs to find a Nash equilibrium of a non-convex function in a high-dimensional parameter space. This scenario may lead to a lack of control during the training phase, exhibiting non-desired side effects such as instability, mode collapse, among others. As a result, many techniques have been proposed to improve the stability training of GANs \cite{salimans2016improved,gulrajani2017improved,miyato2018spectral,durall2019stabilizing}.

\let\thefootnote\relax\footnotetext{Accepted in VISAPP 2021}

Conditioning has risen as one of the key technique in this vein \cite{mirza2014conditional,chen2016infogan,isola2017image,choi2018stargan}, whereby the whole model has granted access to labelled data. In principle, providing supervised information to the discriminator encourages it to behave more stable at training since it is easier to learn a conditional model for each class than for a joint distribution. However, conditioning comes with a price, the necessity for annotated data. The scarcity of labelled data is a major challenge in many deep learning applications which usually suffer from high data acquisition costs. 

Representation learning techniques enable models to discover underlying semantics-rich features in data and disentangle hidden factors of variation. These powerful representations can be independent of the downstream task, leaving the need of labels in the background. In fact, there are fully unsupervised representation learning methods \cite{misra2016shuffle,gidaris2018unsupervised,rao2019continual,milbich2020unsupervised} that automatically extract expressive feature representations from data without any manually labelled annotation. Due to this intrinsic capability, representation learning based on deep neural networks have been becoming a widely used technique to empower other tasks \cite{caron2018deep,oord2018representation,chen2020simple}.

Motivated by the aforementioned challenges, our goal is to show that it is possible to recover the benefits of conditioning by exploiting the advantages that representation learning can offer (see Fig. \ref{fig:teaser}). In particular, we introduce a model that is conditioned on the latent space structure. As a result, our proposed method can generate samples on demand, without access to labeled data at the GAN level. To ensure the correct behaviour, a customized loss is added to the model. Our contributions are as follows.

\begin{itemize}
\item We propose a novel generative adversarial network
conditioned on features from a latent space representation.

\item We introduce a simple yet effective new loss function which incorporates the structure of the latent space.

\item Our experimental results show a neat control on the generated samples. We test the approach on MNIST, CIFAR10 and CelebA datasets.
\end{itemize}

\footnotetext{Standard approach refers to replace the encoded labels with latent code.}

\section{\uppercase{Related Work}}
\subsection{Conditional Generative Adversarial Networks}

Generative image modelling has recently advanced dramatically. State-of-the-art methods are GAN-based models \cite{brock2018large,karras2019style,karras2020analyzing} which are capable of generating high-resolution, diverse samples from complex datasets. However, GANs are extremely sensitive to nearly every aspect of its set-up, from loss function to model architecture. Due to optimization issues and hyper-parameter sensitivity, GANs suffer from tedious instabilities during training.

Conditional GANs have witnessed outstanding progress, rising as one of the key technique to improve stability training and to remove mode collapse phenomena. As a consequence, they have become one of the most widely used approaches for generative modelling of complex datasets such as ImageNet. CGAN \cite{mirza2014conditional} was the first work to introduce conditions on GANs, shortly followed by a flurry of works ever since. There have been many different forms of conditional image generation, including class-based \cite{mirza2014conditional,odena2017conditional,brock2018large}
, image-based \cite{isola2017image,huang2018multimodal,mao2019mode}
, mask- and bounding box-based \cite{hinz2019generating,park2019semantic,durall2020local}, as well as text-based \cite{reed2016generative,xu2018attngan,hong2018inferring}. This intensive research has led to impressive development of a huge variety of techniques, paving the road towards the challenging task of generating more complex scenes.

\subsection{Unsupervised Representation Learning}

In recent years, many unsupervised representation learning methods have been introduced \cite{misra2016shuffle,gidaris2018unsupervised,rao2019continual,milbich2020unsupervised}. The main idea of these methods is to explore easily accessible information, such as temporal or spatial neighbourhood, to design a surrogate supervisory signal to empower the feature learning. Although many traditional approaches such as random projection \cite{li2006very}, manifold learning \cite{hinton2003stochastic} and auto-encoder \cite{vincent2010stacked} have significantly improved feature representations, many of them often suffer  either from being computationally too costly to scale up to large or high-dimensional datasets, or from failing to capture complex class structures mostly due to its underlying data assumption.

On the other hand, a number of recent unsupervised representation learning approaches rely on new self-supervised techniques. These approaches formulate the problem as an annotation free pretext task; they have achieved remarkable results \cite{doersch2015unsupervised,oord2018representation,chen2020simple} and even on GAN-based models as well \cite{chen2019self}. Self-supervision generally involves learning from tasks designed to resemble supervised learning in some way, where labels can be created automatically from the data itself without manual intervention.

\begin{figure*}[t!]
\centering
  \includegraphics[width=0.9\linewidth]{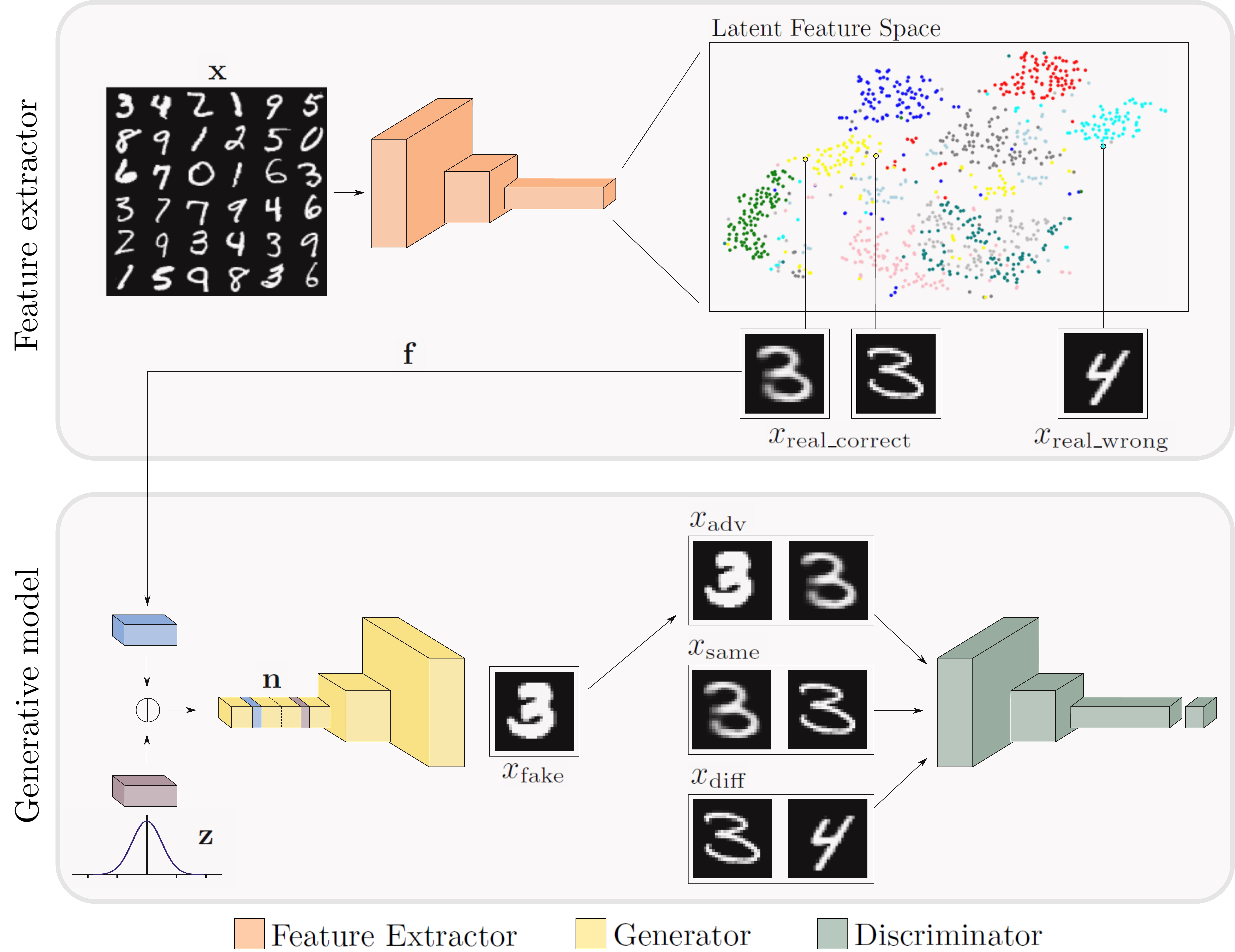}
  \caption{Overview of the processing pipeline of our approach. It contains two main blocks, a feature extraction model and a generative model, more specifically a GAN. This structure allows the generator to incorporate a condition based on the latent space, so that eventually the system can produce images on demand using the latent representation. }
  \label{fig:main}
\end{figure*}

\section{\uppercase{Method}}

In this section we describe our approach in detail. First, we present our representation learning set-up together with its sampling algorithm. Then, we introduce a new loss function capable of exploiting the structural properties from the latent space. Finally, we have a look at the adversarial framework for training a model in an end-to-end fashion. Fig. \ref{fig:main} gives an overview of the main pipeline and its  components.

\subsection{Representation Learning}

The goal of representation learning or feature learning is to find an appropriate representation of data in order to perform a machine learning task.\\

\noindent \textbf{Generating Latent Space.}
The latent space must contain all the important information needed to represent reliably the original data points in a simplified and compressed space. Similar to \cite{aspiras2019active}, in our work we also try to exploit the latent space. In particular, we rely on existing topologies that can capture a high level of abstraction. Hence, we mainly focus on integrating these data descriptors and on evaluating their usability and impact. For this reason, We count on several set-ups where we can sample informative features of different qualities, i.e. level of clustering in the latent spaces.

Our feature extractor block $E$ is a convolutional-based model for classification tasks. Inspired by \cite{caron2018deep}, to extract the features we do not use the classifier output logits, but the feature maps from an intermediate convolutional layer. We refer to these hidden spaces as latent space.\\

\noindent \textbf{Sampling from Latent Space.}
Assuming that the feature extractor is able to produce a structured latent space, e.g. semi-clustered features, we can start sampling observations that will be fed to our GAN afterwards. The procedure to create sampling batches is described in Algorithm \ref{alg:sampling}.

\begin{algorithm}
	\caption{Creating batches for training the GAN.}
	\label{alg:sampling}
	\begin{algorithmic}[1]	
  	\STATE Sample a batch of images $\mathrm{\mathbf{x}}$
  	\STATE Extract features from it $\mathbf{f} = E(\mathbf{x})$ 
  	\STATE Compute distance between all features $\mathbf{D} = ||\mathbf{f}||_{1}$
  	\FOR {$x_i$ in $\mathrm{\mathbf{x}}$}
  	
  	\STATE Select $x_i$ and sort the rest according to their distance $d(x_i)$
  	\STATE Select nearest neighbour from $x_i$, i.e. $d_\mathrm{min}(x_i)$
  	\STATE Select the farthest neighbour from $x_i$, i.e. $d_\mathrm{max}(x_i)$
  
  	\ENDFOR
 	\end{algorithmic} 
\end{algorithm}

\subsection{Loss Function}

\noindent \textbf{Minimax Loss.}
A GAN architecture is comprised of two parts, a discriminator $D$ and a generator $G$. While the discriminator trains directly on real and generated images, the generator trains via the discriminator model. They should therefore use loss functions that reflect the distance between the distribution of the data generated $p_{\mathrm{z}}$ and the distribution of the real data $p_{\mathrm{data}}$. Minimax loss is by default the candidate to carry on with this task and it is defined as

\begin{align}
\begin{split}
	\min_{G} \max_{D} \mathcal{L}(D,G) =\,  \mathbb{E}_{\mathrm{\mathbf{x}} \sim p_{\mathrm{data}}} \left[ \log \left(D(x)\right) \right] + \cr \mathbb{E}_{z \sim p_{z}}[\log(1-D(G(z)))].
\end{split}
\label{formula:main}
\end{align}
\vspace{1px}

\noindent \textbf{Triple Coupled Loss.}
In the vanilla minimax loss the discriminator expects batches of individual images. This means that there is a unique mapping between input image and output, where each input is evaluated and then classified as real or fake. Despite being a functional loss term, if we hold to that closed formulation, we cannot leverage alternatives such as conditional features or combinatorial inputs i.e. input is not any longer only a single image but a few of them.

We introduce a loss function coined triple coupled loss that incorporates combinatorial inputs acting as a semi-conditional mechanism. The approach lies on the idea of exploiting similitudes and differences between images. In fact, similar approaches have been already successfully implemented in other works \cite{chongxuan2017triple,sanchez2018triple,ho2020learning}. In our implementation, the new discriminator takes couples of images as input and classify them as true or false. Unlike minimax case, now we have two degrees of freedom (two inputs) to take advantage of. Therefore, we produce different scenarios to further enhance the capabilities of our discriminator, so that it can also be conditioned in an indirect manner by the latent representation space. We can distinguish three different coupled case scenarios and their corresponding losses

\begin{align}
\begin{split}
	x_{\mathrm{adv}} = [x_{\mathrm{real}\_\mathrm{correct}},x_{\mathrm{fake}}] \,\longrightarrow\, \mathcal{L}_{\mathrm{adv}}\\
	\mathcal{L}_{\mathrm{adv}} = \mathbb{E}_{\mathrm{\mathbf{x}} \sim ( p_{\mathrm{data}} \,\cup\, p_{\mathrm{z}} ) } \left[ \log \left(1-D(x_{\mathrm{adv}})\right) \right]
\end{split}
\end{align}
\begin{align}
\begin{split}
	x_{\mathrm{same}} = [x_{\mathrm{real}\_\mathrm{correct}},x_{\mathrm{real}\_\mathrm{correct}}] \,\longrightarrow\, \mathcal{L}_{\mathrm{same}}\\
	\mathcal{L}_{\mathrm{same}} = \mathbb{E}_{\mathrm{\mathbf{x}} \sim p_{\mathrm{data}}} \left[ \log \left(D(x_{\mathrm{same}})\right) \right]
\end{split}
\end{align}
\begin{align}
\begin{split}
	x_{\mathrm{diff}} = [x_{\mathrm{real}\_\mathrm{correct}},x_{\mathrm{real}\_\mathrm{wrong}}] \,\longrightarrow\, \mathcal{L}_{\mathrm{diff}}\\
	\mathcal{L}_{\mathrm{diff}} = \mathbb{E}_{\mathrm{\mathbf{x}} \sim p_{\mathrm{data}}} \left[ \log \left(1-D(x_{\mathrm{diff}})\right) \right].
\end{split}
\end{align}
\vspace{1px}

We first have $x_{\mathrm{adv}}$ case which is the combination of one generated image ($x_{\mathrm{fake}}$) and one real that belongs to the target class ($x_{\mathrm{real}\_\mathrm{correct}}$). Then, we have $x_{\mathrm{same}}$ with two different "real correct" samples. Finally, the last case is $x_{\mathrm{diff}}$ which combines one "real correct" and one "real wrong". The latter term is a real image from a different class, i.e. not target class ($x_{\mathrm{real}\_\mathrm{wrong}}$).

In order to incorporate the triple coupled loss, we need to reformulated the Formula \ref{formula:main} adding the aforementioned three case scenarios. As a result, the new objective loss is rewritten as follows

\begin{align}
\begin{split}
	\min_{G} \max_{D} \mathcal{L}(D,G) =\,& \lambda_{\mathrm{a}} \mathcal{L}_{\mathrm{adv}} \,+\, \lambda_{\mathrm{s}} \mathcal{L}_{\mathrm{same}} \,+\, \lambda_{\mathrm{d}} \mathcal{L}_{\mathrm{diff}}
\end{split}
\end{align}

\noindent  where $\lambda$s are the weighting coefficients.

\subsection{Training on Conditioned Latent Feature Spaces}

Our approach is divided into two distinguishable elements, the feature extractor $E$ and the generative model. With the integration of these two components into an embedded system, our model can produce samples on demand without label information.\\

\noindent \textbf{Dynamics of Training.}
Given an input batch $\mathrm{\mathbf{x}}$, the feature extractor produces the latent code $\mathrm{\mathbf{f}}$. Then, we generate a vector of random noise $\mathrm{\mathbf{z}}$ (e.g. Gaussian) and we attach to it the $\mathrm{\mathbf{f}}$, creating in this way the input for our generator $\mathrm{\mathbf{n}}$ (see Fig. \ref{fig:main}).

The expected behaviour from our generator should be similar to CGAN, where the generator needs to learn a twofold task. On the one hand, it has to learn to generate realistic images by approximating the real data distribution as much as possible. On the other hand, these synthetic images need to be conditioned consistently on $\mathrm{\mathbf{f}}$, so that later can be controlled. For example, when two similar\footnote[2]{Similarity is measured by $l_1$ distance as described in Algorithm \ref{alg:sampling}.} latent codes are fed into the model, this should produce two similar output images belonging to the same class.

\begin{figure*}[t!]
\centering
  \includegraphics[width=\linewidth]{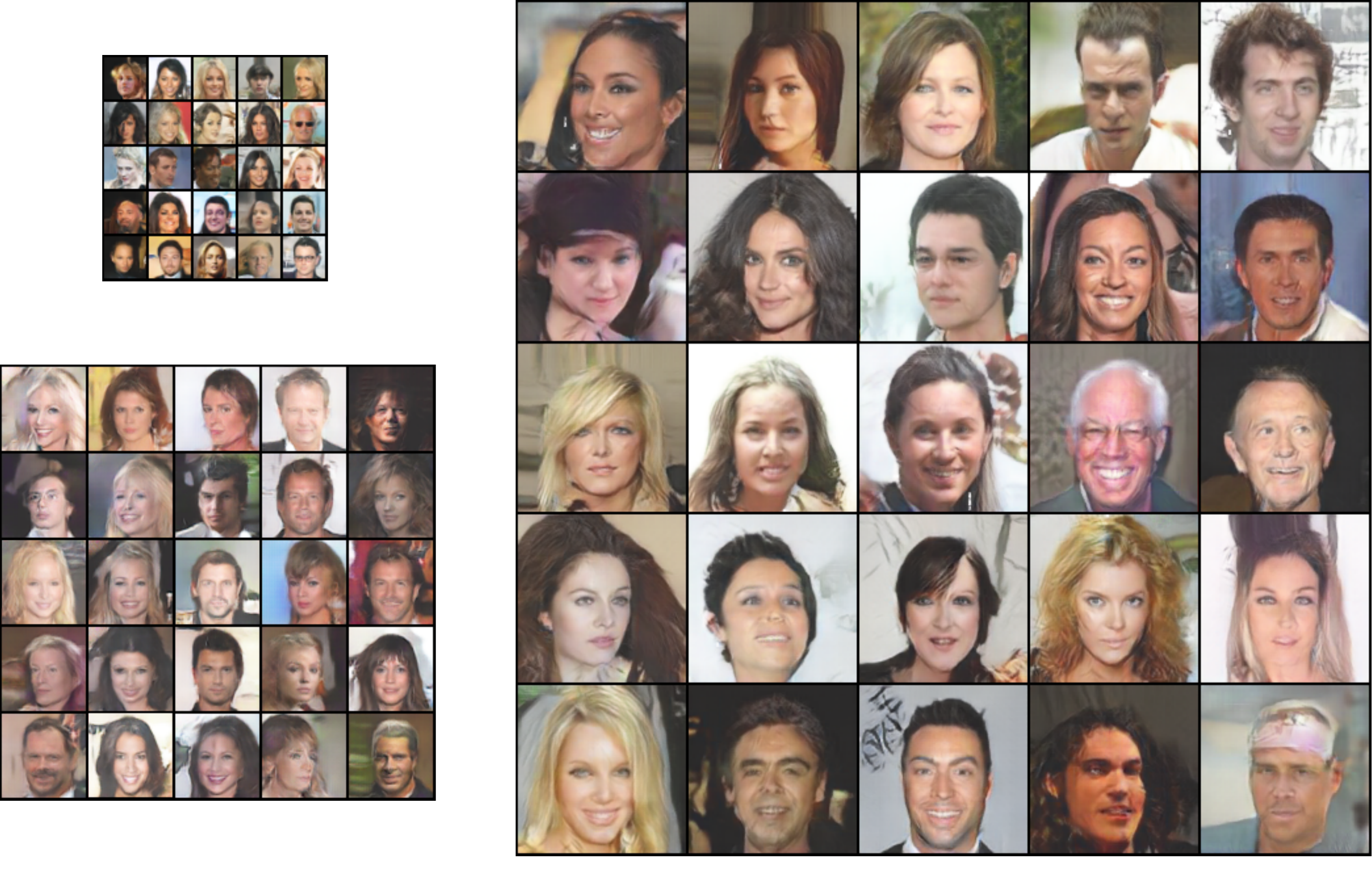}
  \caption{Random generated samples of 32x32, 64x64 and 128x128 resolutions.}
  \label{fig:scale}
\end{figure*}

The discriminator, however, has a remarkable difference with CGAN when it comes to training. While CGAN employs latent codes to condition directly the outcome results, our method uses a semi-conditional mechanism through the coupled inputs. As it is explained in the upper section, the discriminator employs the triplet coupled loss which enforces to respect the latent space structure and binding in this way the output with the conditional information. Algorithm \ref{alg:training} describes the training scheme.

\begin{algorithm}
	\caption{Training GAN model.}
	\label{alg:training}
	\begin{algorithmic}[1]	
\STATE Require: $n_{\mathrm{iter}}$, the number of iterations. $n$, the number of iterations of the generator per discriminator iteration. $\lambda$'s, the weighting coefficients. $\theta_{\mathrm{gen}}$, generator's parameters. $\theta_{\mathrm{disc}}$, discriminator's parameters. 

  \FOR {$i < n_{\mathrm{iter}}$}
  
  \STATE Sample batch using Algorithm 1
  
  \STATE \# Train generator $G$
  \STATE $\mathcal{L}_{\mathrm{gen}} =  \mathcal{L}_{\mathrm{adv}}$
  \STATE $\theta_{\mathrm{gen}} \leftarrow \theta_{\mathrm{gen}} +  \nabla \mathcal{L}_{\mathrm{gen}}$
  \IF {$mod(i,n) = 0$}
  \STATE \# Train discriminator $D$
  \STATE $\mathcal{L}_{\mathrm{disc}} = \lambda_{\mathrm{a}} \mathcal{L}_{\mathrm{adv}} \,+\, \lambda_{\mathrm{s}} \mathcal{L}_{\mathrm{same}} \,+\, \lambda_{\mathrm{d}} \mathcal{L}_{\mathrm{diff}}$
  \STATE $\theta_{\mathrm{disc}} \leftarrow \theta_{\mathrm{disc}} + \nabla \mathcal{L}_{\mathrm{disc}}$ 
  \ENDIF
  \ENDFOR
 	\end{algorithmic} 
\end{algorithm}

\section{\uppercase{Experiments}}

In this section, we show results for a series of experiments evaluating the effectiveness of our approach. We first give a detailed introduction of the experimental set-up. Then, we analyse the response of our model under different scenarios and we investigate the role that plays the structure of the latent space and its robustness. Finally, we check the impact of our customized loss function though an ablation study.

\subsection{Experimental Set-up}

We conduct a set of experiments on MNIST \cite{lecun1998gradient}, CIFAR10 and CelebA \cite{liu2015faceattributes} datasets. For each one, we use an individual classifier to ensure certain structural properties on our latent space. Next, we extract feature from one intermediate layer, and feed them into our generative model.\\

\noindent \textbf{MNIST.}
The experiments carried on MNIST are fully\textbf{ unsupervised} since we do not require any label information.  We choose to deploy an untrained AlexNet model \cite{krizhevsky2012imagenet} as feature extractor. As shown in \cite{caron2018deep}, AlexNet offers an out-of-the-box clustered space at certain intermediate layer without any need of training. Hence, we extract there the features and no extra processing step is involved.\\

\noindent \textbf{CIFAR10.}
Despite the fact that CIFAR10 is fairly close to MNIST in terms of amount of samples and classes (10 in both cases), it is indeed a much more complex dataset. As a result, in this case we need to train a feature extractor to achieve a structured latent space. Inspired by the unsupervised representation learning method \cite{gidaris2018unsupervised}, we build a classifier which reaches similar accuracy.\\

\noindent \textbf{CelebA.}
Different from the previous datasets, CelebA contains only "one class" of images. In particular, this datatset is an extensive collection of faces. However, each sample can potentially contain up to 20 different attributes. So, in our experiments we build different scenarios by splitting the dataset into different classes according to their attributes, e.g. \textit{man} and \textit{woman}. Moreover,  we also test our approach on different resolutions, since the size of the images of CelebA is larger. Similarly to CIFAR10 case, we need to train again a feature extractor.

\begin{figure*}[t!]
\begin{subfigure}{0.49\linewidth}
	\centering
	\includegraphics[width=\linewidth]{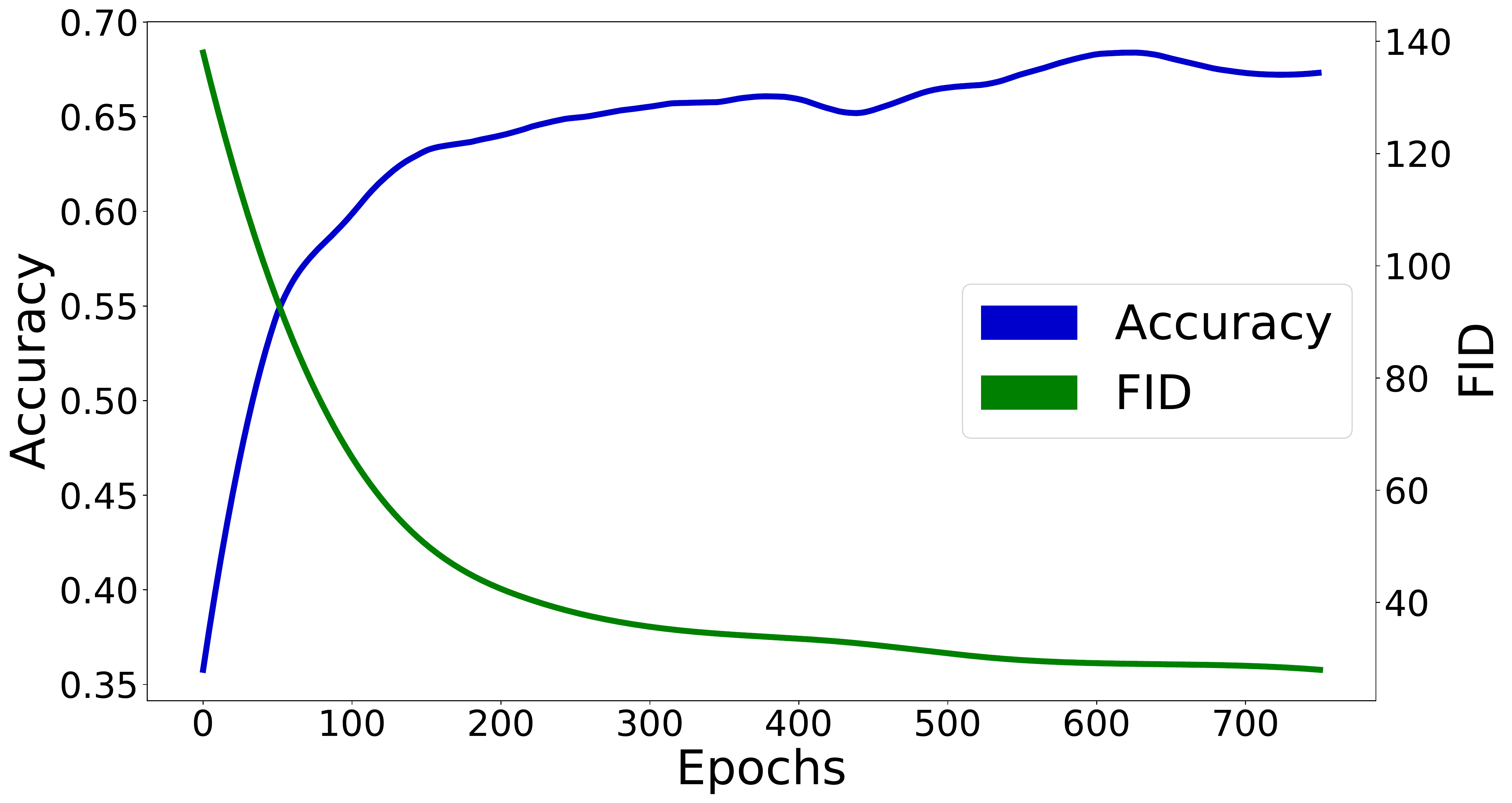}
	\caption{CIFAR10.}
	\label{fig:cifar_curve}
\end{subfigure}
\begin{subfigure}{0.49\linewidth}
	\centering
	\includegraphics[width=\linewidth]{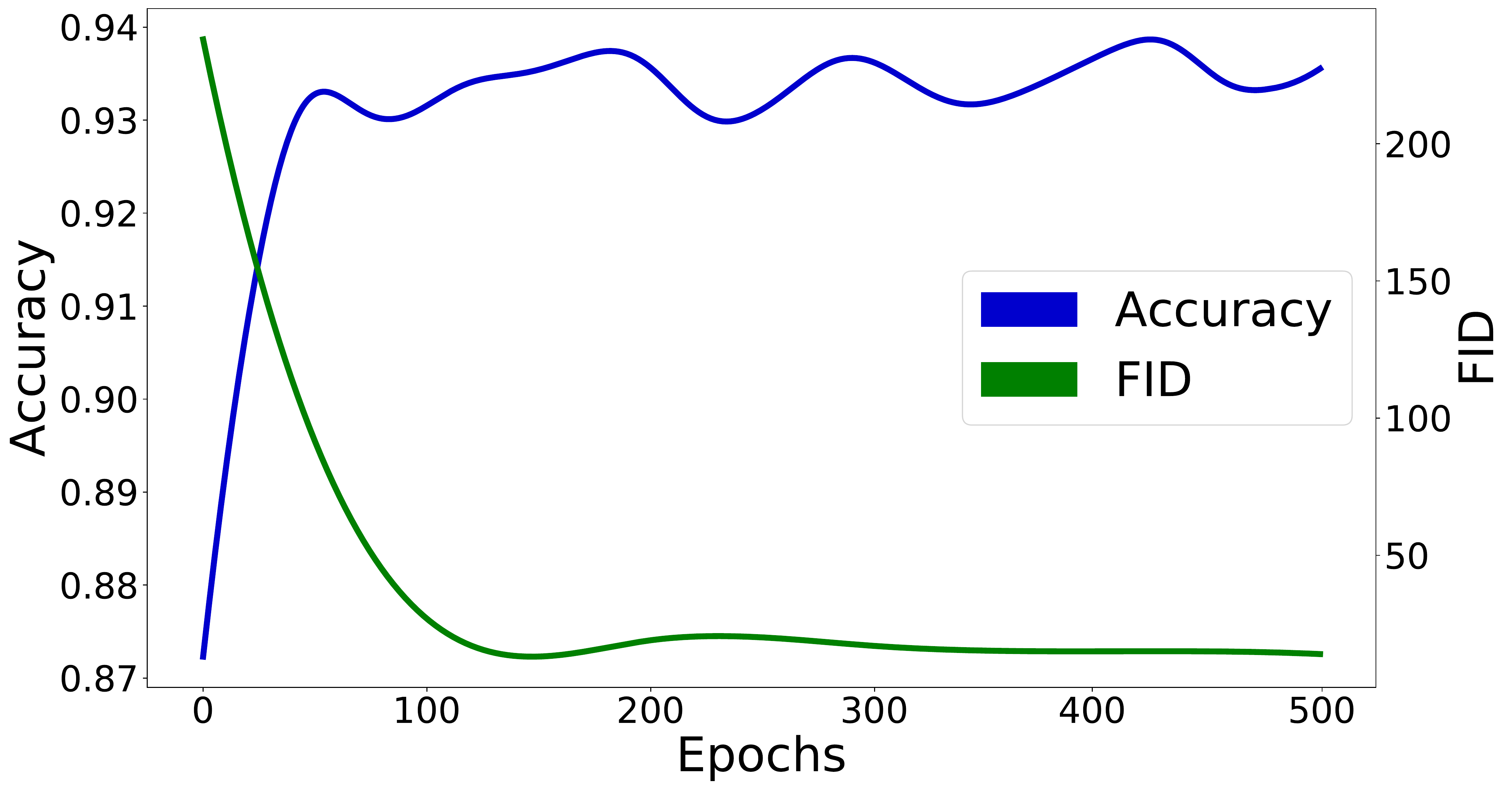}
	\caption{CelebA.}
	\label{fig:cifar_curve}
\end{subfigure}
\caption{FID and accuracy learning curves on CIFAR10 and CelebA.}
\label{fig:celeba_evol}
\end{figure*}

\begin{table}[t!]
 \centering
 \caption{Validation results in MNIST, CIFAR10 and CelebA.}
	\begin{tabular}{cccc}
	 	\hline
    	\textbf{MNIST} & IS & FID & Accuracy \\
   		\hline
    		baseline & 9.63 & - & - \\
    		ours & 9.78 & - & 72\% \\
 		\hline
    	\textbf{CIFAR10} & IS & FID & Accuracy \\
   		\hline
    		baseline & 7.2 & 28.76 & - \\
    		ours & 7.0 & 29.71 & 68\% \\
    	\hline
 		\textbf{CelebA} & IS & FID & Accuracy \\
   		\hline
    		baseline & 2.3 & 18.56 & - \\
    		ours (32) & 2.5 & 11.10 & 90\% \\
    		ours (64) & 2.7 & 13.69 & 94\% \\
    		ours (128) & 2.65 & 37.59 & 94\% \\
    	\hline
    \end{tabular} 
    \label{table:metric}
\end{table}

\subsection{Evaluation Results}
We compare the baseline model based on 
Spectral Normalization for Generative Adversarial Networks (SNGAN) \cite{miyato2018spectral} to our approach that incorporates the latent code and the coupled input on top of it. The rest of the topology remains unchanged.\footnote[3]{In CelebA, we add one and two layers into the model to be able to produce samples with resolution of 64x64 and 128x128, respectively.} We do not use CGAN architecture since our framework is not conditioned on labels. Therefore, we take an unsupervised model as a baseline. In particular, we choose SNGAN because it is a simple yet stable model that allows to control the changes applied on the system. Besides, it generates appealing results having a competitive metric scores.

To evaluate generated samples, we report standard qualitative scores on the Frechet Inception Distance (FID) and the Inception Score (IS) metrics. Furthermore, we provide the accuracy scores that eventually quantize the success of the system. We compute this score using a classifier trained on the real data that guarantees that the metric correctly assesses the percentage of generated samples that coincide with the class of the latent code. For instance, if the latent code belongs to a \textit{cat}, the generator should produce a \textit{cat}. 

Table \ref{table:metric} compares scores for each metric. We observe how our model performs fairly similar to the baseline independently of the scenario. Only when we ask for a 128x128 output resolution, the FID score increases substantially. We hypothesize that this break happens due to a model architecture issue since the baseline is initially designed for 32x32 images (see Fig. \ref{fig:scale}). Fig. \ref{fig:celeba_evol} plots FID and accuracy training curves on CIFAR10 and CelebA datasets, and confirms that our approach exhibits a strong correlation between the both metrics. A better FID score (low value) means always a higher accuracy score.

\begin{figure*}[t!]
\centering
  \includegraphics[width=\linewidth]{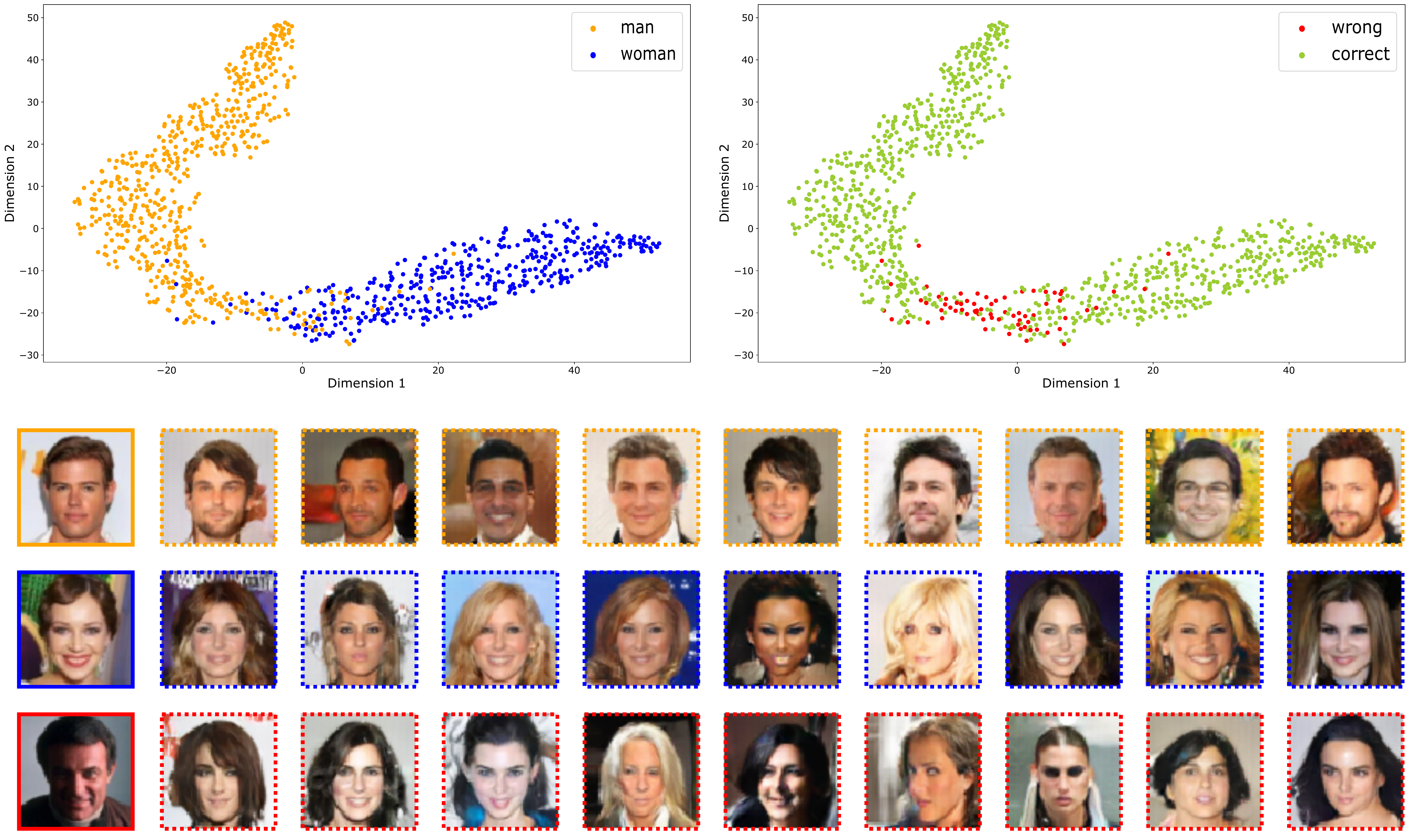}
  \caption{Visualization of border effect on CelebA for the classes \textit{man} and \textit{woman}. Upper-left: t-SNE of extracted features regarding to their classes. Upper-right: t-SNE of extracted features regarding to their capacity of conditioning the output i.e. accuracy. Bottom: Random samples from different latent codes, where the solid frame belong to the real images and the dashed frames the generated. The code color is consistent within the whole figure.}
  \label{fig:tsne_celeba2}
\end{figure*}

It is important to notice that baseline models do not have accuracy score since they cannot choose the class of the output. On the other hand, regarding our approach, the accuracy for both MNIST and CIFAR is around 70\%, and more than 90\% for CelebA. This gap is directly related with the quality of the latent space. In other words, the more clustered the latent space is, the higher accuracy our model can have. In this case, CelebA is evaluated in a scenario with only two classes \textit{man} and \textit{women}, as a consequence the latent spaces is simpler. As a rule of thumb, an increase of classes will often lead to a more tangled latent space making the problem harder. The main reason for that are those samples located on the borders. We refer to this phenomenon as border effect and it is shown in Fig. \ref{fig:tsne_celeba2}. As it is expected, we observe how the samples that lie between the two blobs have usually a higher failure rate (colored in red).

\subsection{Impact of the Latent Space Structure}
The structure of latent space plays an important role and has a direct effect in the accuracy performance. This is mainly due to the nature of the triple coupled loss. This term relies on having at least a semi-clustered feature space to sample from. Hence, those latent spaces with almost no structure will build many false couples in training time and resulting in bad performance. Notice that the generator model does not use label information directly, but through the extracted features.

\begin{table*}[t!]
 \centering
 \caption{Statistics of the latent space's structure for different scenarios. }
 	\begin{tabular}{cccccc}
 		\hline
 		\textbf{MNIST} & Classes &  Accuracy & $\;1^{\mathrm{st}}$ neighbour & $\;2^{\mathrm{nd}}$ neighbour& $\;5^{\mathrm{th}}$ neighbour\\
   		\hline
    		baseline  & 10 & - & 10\% & 10\% & 10\%\\
    		ours  & 10 & 72\% & 89\% & 84\% & 78\%\\
    	\hline
    	\textbf{CIFAR10} & Classes & Accuracy & $\;1^{\mathrm{st}}$ neighbour & $\;2^{\mathrm{nd}}$ neighbour& $\;5^{\mathrm{th}}$ neighbour\\
   		\hline
    		baseline  & 10 & - & 10\% & 10\% & 10\%\\
    		ours  & 10 & 68\% & 70\% & 68\% & 65\% \\
    	\hline
 		\textbf{CelebA}  & Classes & Accuracy & $\;1^{\mathrm{st}}$ neighbour & $\;2^{\mathrm{nd}}$ neighbour& $\;5^{\mathrm{th}}$ neighbour\\
   		\hline
    		baseline  & 2 & - & 50\% & 50\% & 50\%\\
    		ours (32)  & 2 & 90\% & 88\% & 88\% & 86\%\\
    		ours (64)  & 2 & 94\% & 95\% & 94\% & 93\%\\
    		\hline
    		baseline  & 5 & - & 20\% & 20\% & 20\%\\
    		ours (32)  & 5 & 80\% & 90\% & 89\% & 87\%\\
    		ours (64)  & 5 & 78\% & 96\% & 95\% & 92\%\\
    	\hline
    \end{tabular} 
    \label{table:statistic}
\end{table*}

\begin{figure*}[t!]
\begin{subfigure}{0.49\linewidth}
	\centering
	\includegraphics[width=\linewidth]{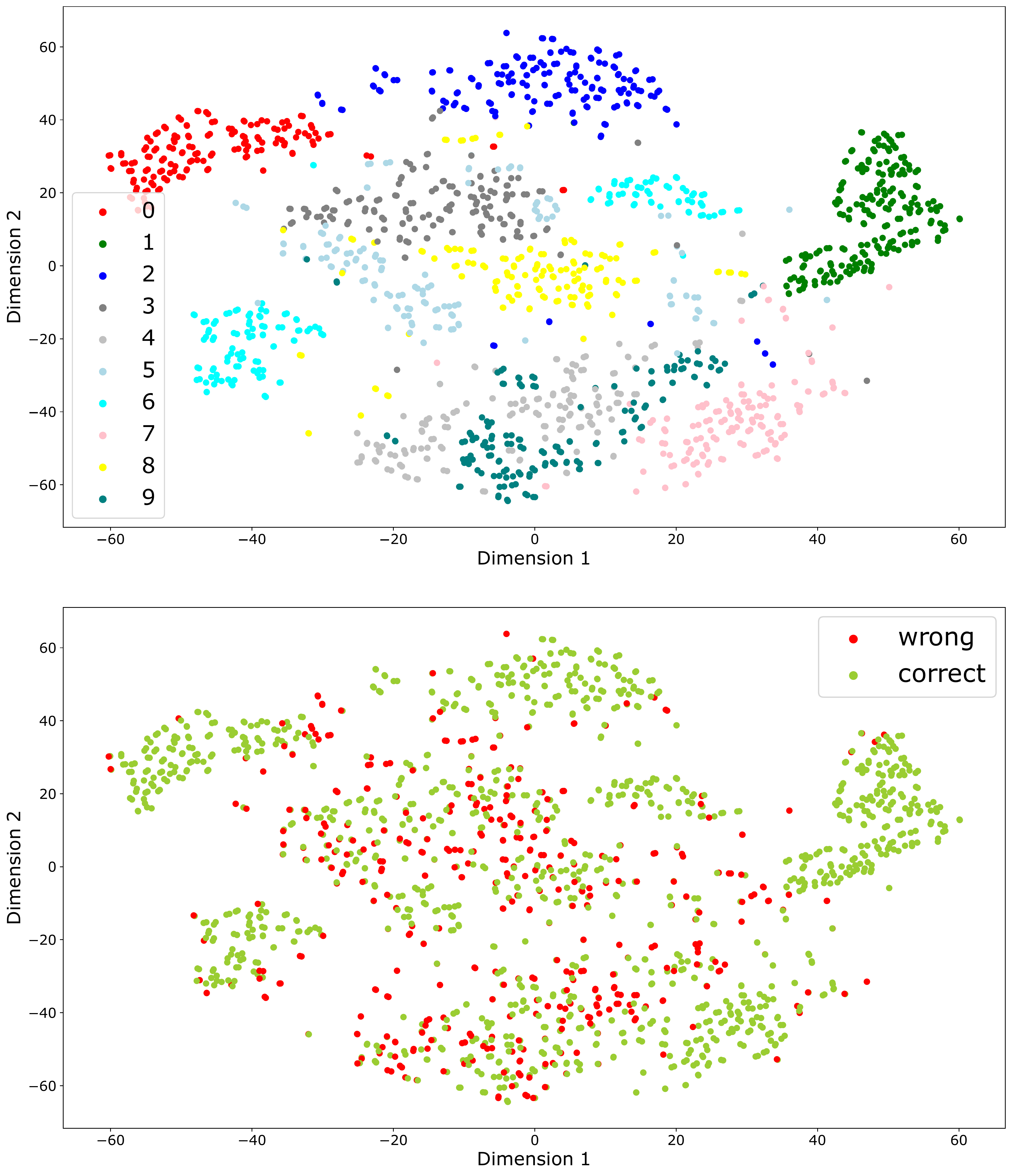}
	\caption{Semi-clustered latent space.}
	\label{fig:mnist_normal}
\end{subfigure}
\begin{subfigure}{0.49\linewidth}
	\centering
	\includegraphics[width=\linewidth]{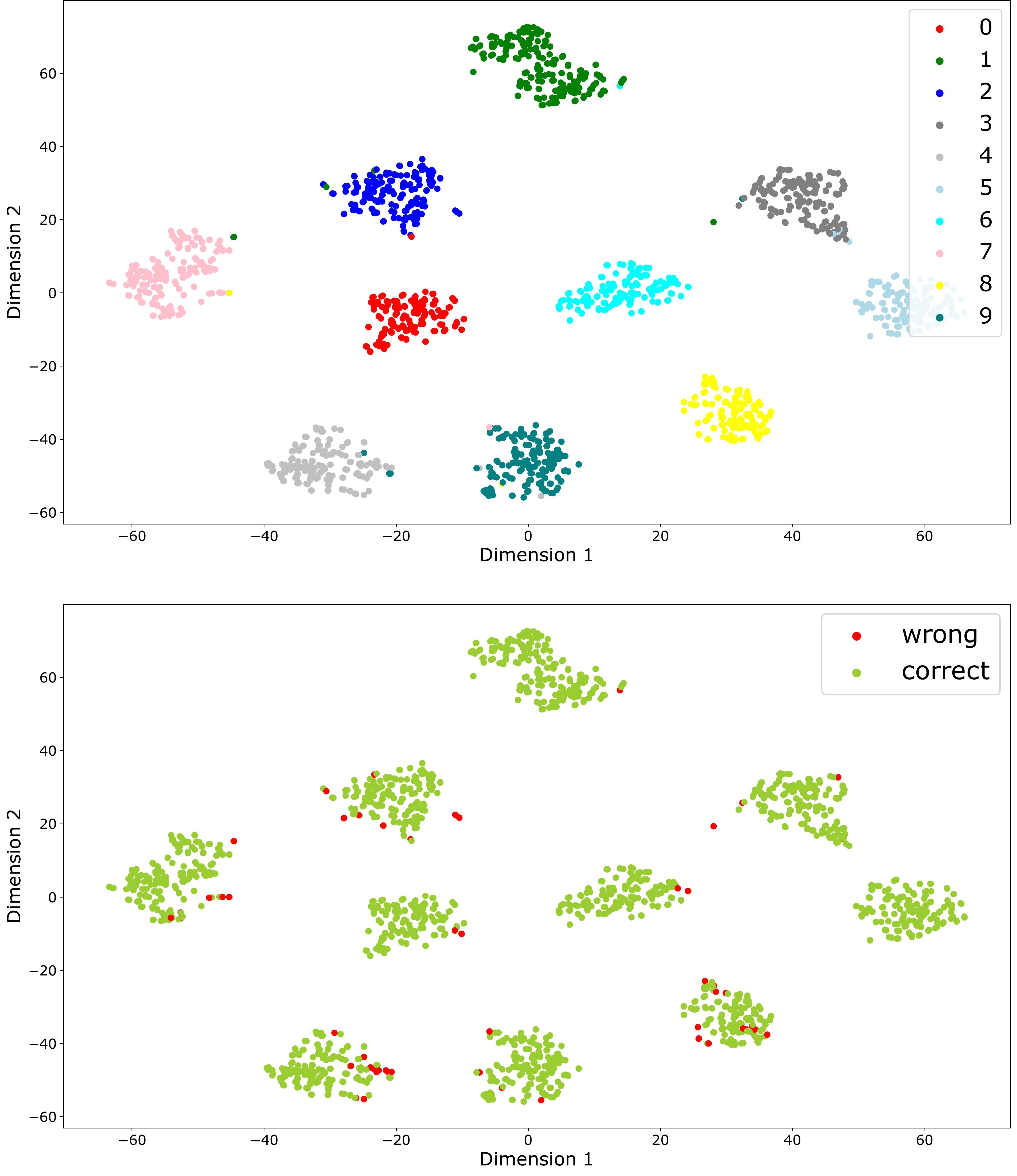}
	\caption{Full-clustered latent space.}
	\label{fig:mnist_cluster}
\end{subfigure}
\caption{t-SNE visualizations from two different latent spaces on MNIST. First row displays the classes and second row the accuracy from our approach.}
\label{fig:mnist}
\end{figure*}

We run the evaluations on MNIST, CIFAR10 and CelebA datasets as in the previous section. However, in CelebA's case there are now two different set-ups. One based on gender (\textit{man} and \textit{woman}), and a second one based on hair (\textit{blond}, \textit{black}, \textit{brown}, \textit{gray} and \textit{bold}). In order to study the impact of the latent space structure, we need to determine how clustered our space is. Therefore, we compute a set of statistics (see Table \ref{table:statistic}) that are useful to estimate the initial conditions of the latent structure, and consequently find out the boundaries that our system might not overcome. For example, our model on CIFAR10 reports 70\% on $\;1^{\mathrm{st}}$ neighbour. This value indicates that if we take one random sample from our latent space, 70\% of the time its nearest neighbour will belong to the same class. Empirically, we observe the causal effect that the structure of latent space has on the accuracy results. The more clustered, i.e. higher neighbours scores, the better the accuracy. In other words, neighbourhood information helps to understand the upper-bounds fixed by the latent space.

Fig. \ref{fig:mnist} compares two identical set-ups with different latent spaces. On the one hand, we have the semi-clustered space produced by an untrained AlexNet. This scenario achieves good accuracy scores despite the border effect. On the other hand, we have an extreme case with a fully-clustered space. As expected, all the scores are dramatically improved at the cost of having a perfect space.

\subsection{Robustness of the Latent Space}

\begin{figure}[t]
\centering
  \includegraphics[width=\linewidth]{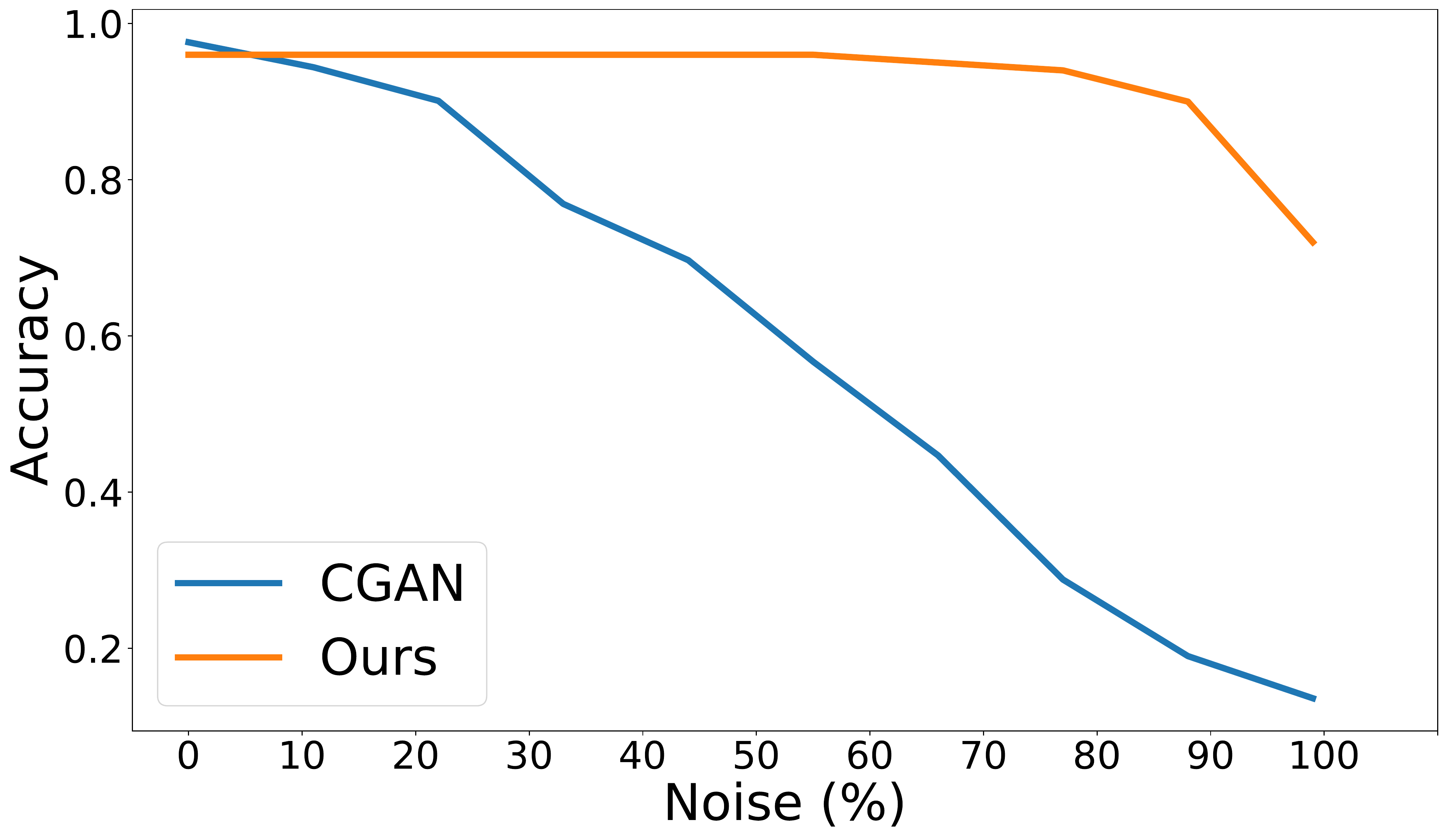}
  \caption{Robustness evaluation on MNIST using accuracy curves.}
  \label{fig:cgan}
\end{figure}

In this section, we analyse how our approach behaves when we introduce noisy labels, and we compare it to CGAN performance. This analysis allows us to quantify how robust our system is. 
We start the experiments having a set-up free of noise.\footnote[4]{Notice that for this experiment we take a feature extractor and we train it from scratch each time that we change the percentage of noise.} Then, we gradually increase the amount of noise by introducing noisy labels. Fig. \ref{fig:cgan} shows the accuracy curves evolution for both cases. We observe how CGAN has almost a perfect lineal relationship between noise and accuracy. Every time that noise increases, the accuracy decreases in a similar proportion. This demonstrate the necessity of CGAN of labels to produce the desired output and the incapacity to deal with noise. Therefore, its robustness against noise very is limited. On the other hand, our approach shows a more robust behaviour. In this case, there is not lineal relationship, and the system is able to maintain the accuracy score independently of the level of noise. Only a notable decrease happens when the percentage of noisy labels surpasses the barrier of 90\%.

\begin{figure*}
\begin{subfigure}{0.49\linewidth}
	\centering
	\includegraphics[width=\linewidth]{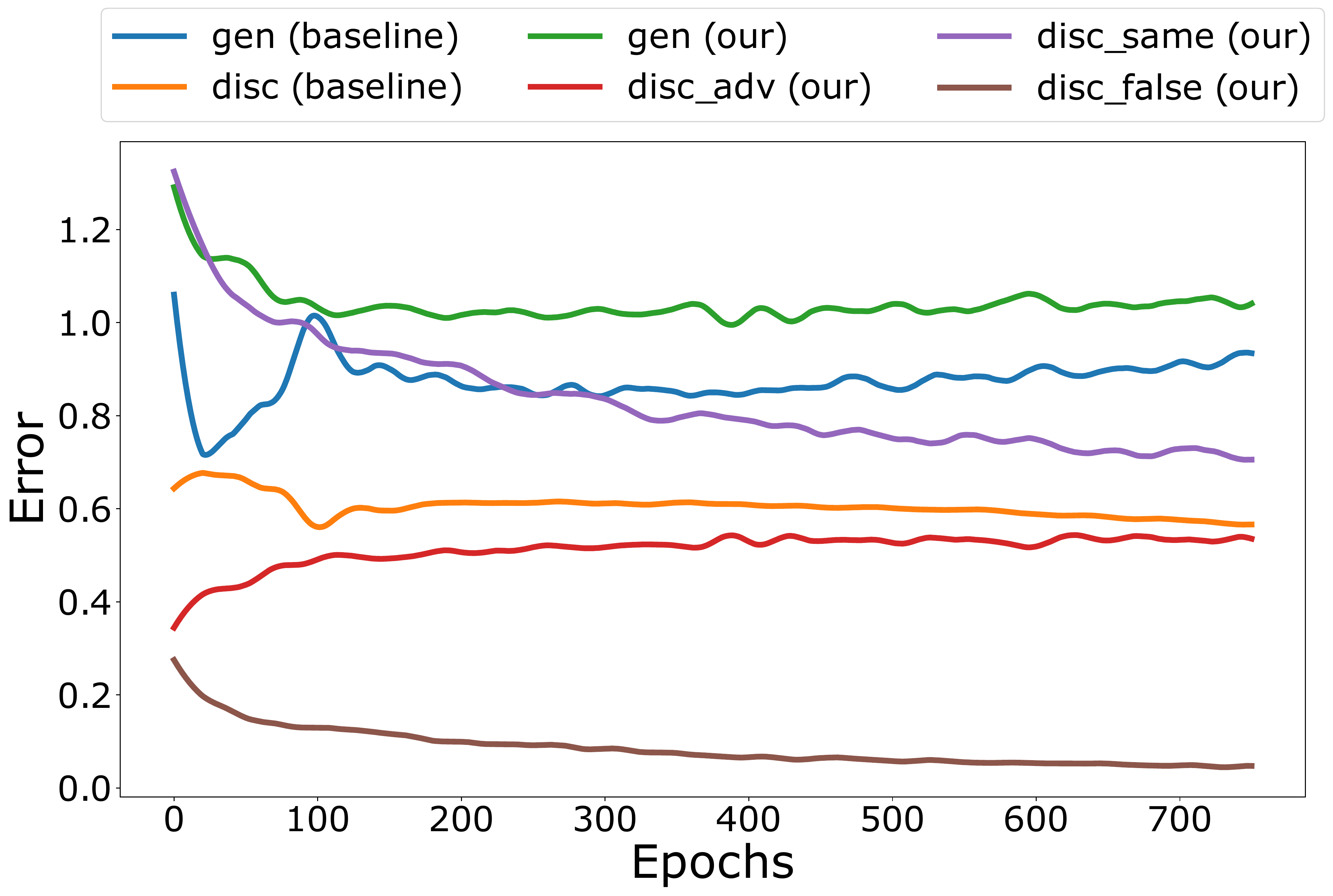}
	\caption{Losses evolution.}
	\label{fig:cifar_loss}
\end{subfigure}
\begin{subfigure}{0.49\linewidth}
	\centering
	\includegraphics[width=\linewidth]{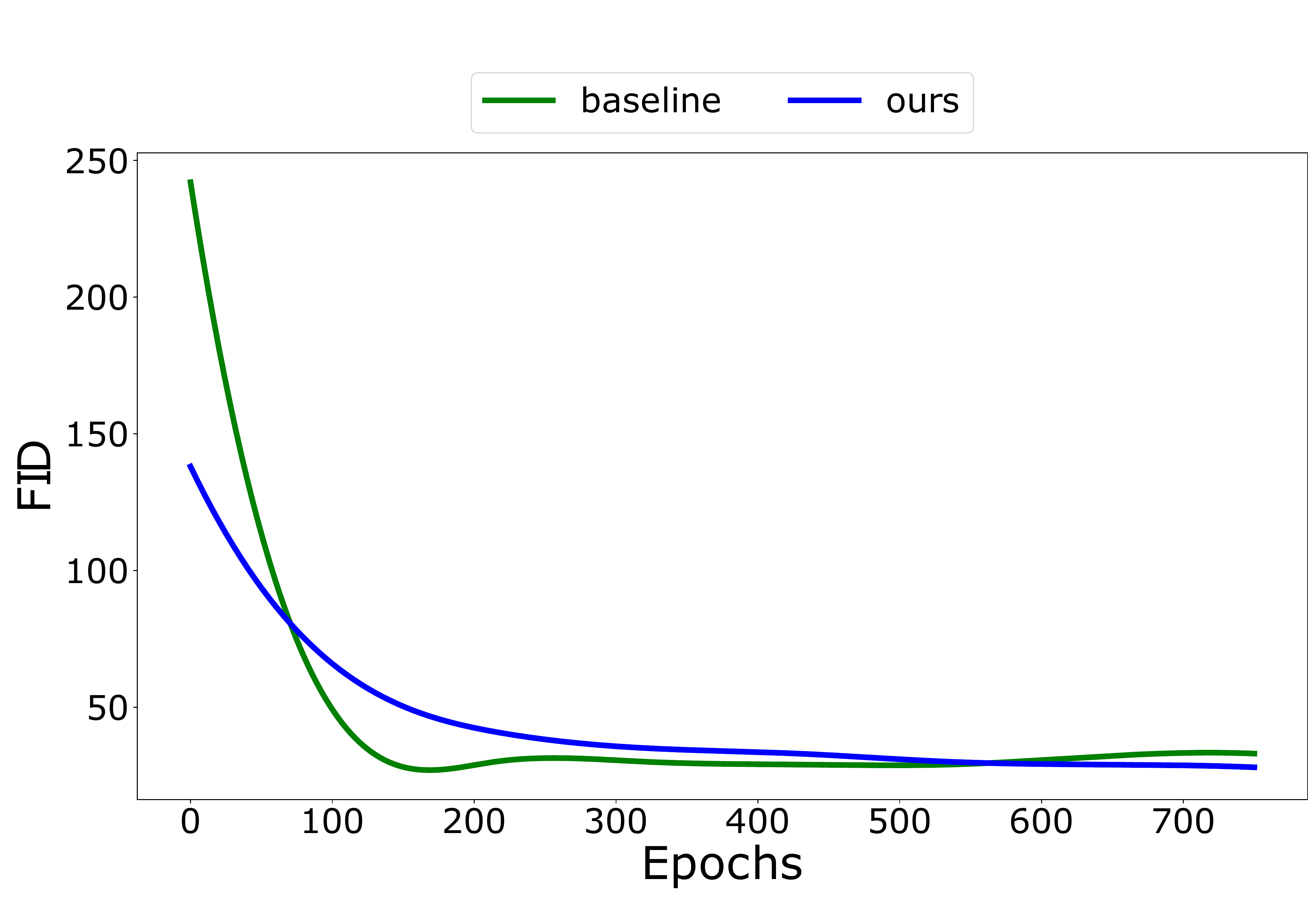}
	\caption{FID evolution.}
	\label{fig:cifar_comp}
\end{subfigure}
\caption{Comparison between the baseline and our approach on CIFAR10.}
\label{fig:cifar}
\end{figure*}

\begin{table*}
 \centering
 \caption{Quantitative results of the ablation study on CIFAR10.}
    \begin{tabular}{ccccccc}
 		\hline
 		& $\mathcal{L}_{\mathrm{minimax}}$ &  $\mathcal{L}_{\mathrm{adv}}$ &  $\mathcal{L}_{\mathrm{same}}$ &  $\mathcal{L}_{\mathrm{diff}}$ & Convergence & Accuracy\\
    	\hline
    	baseline &  \checkmark & & & & \checkmark & 8\% \\
    	prototype A &  & \checkmark & & &  &  - \\
    	prototype B &  & \checkmark & \checkmark & & \checkmark & 58\%\\
    	prototype C &  & \checkmark & & \checkmark &  & -\\
    	ours &  & \checkmark & \checkmark & \checkmark & \checkmark &68\%\\
    	\hline
    \end{tabular}
    \label{table:ablation}
\end{table*}

\subsection{Analysis of Triple Coupled Loss}
The triple coupled loss is designed to exploit the structure of the latent space, so that the generative model learns how to produce samples on demand, i.e. based on the extracted features. Fig. \ref{fig:cifar} shows the itemized losses and FID learning curves for minimax loss (baseline) and for triple coupled loss (ours). We can confirm a similar behaviour between our model and the baseline, having as a side-effect an increase of convergence training time. In exchange for this delay, the proposed system has control over the outputs through the extracted features.

\section{\uppercase{Ablation Study}}

In this section, we quantitatively evaluate the impact of removing or replacing parts of the triple coupled loss. We do not only illustrate the benefits of the proposed loss compared to minimax loss, but also present a detailed evaluation  of our approach. Table \ref{table:ablation} presents how the loss function behaves when we modify its  components. First, we check whether the sub-optimal loss leads the system to convergence, i.e. it is able to generate realistic images. And second, for those functions that have the capacity of generating, we check their accuracy score. Notice that ideally we would achieve 100$\%$ which means producing the desired output all the time.

Based on the empirical results from the previous table, we can see the importance of each term in the loss function. In particular, we observe how $\mathcal{L}_{\mathrm{same}}$ is essential to achieve convergence, and how the combination of all three terms brings the best result.

\section{\uppercase{Conclusions}}

Motivated by the desire to condition GANs without using label information, in this work, we propose an unsupervised framework that exploits the latent space structure to produce samples on demand. In order to be able to incorporate the features from the given space, we introduce a new loss function. Our experimental results show the effectiveness of the approach on different scenarios and its robustness against noisy labels.

We believe the line of this work opens new avenues for feature research, trying to combined different unsupervised set-ups with GANs. We hope this approach can pave the way towards high quality, fully unsupervised, generative models.

\bibliographystyle{apalike}
{\small
\bibliography{example}}

\end{document}